\documentclass[sigconf]{aamas}

\setcopyright{ifaamas}
\acmConference[AAMAS '26]{Proc.\@ of the 25th International Conference
on Autonomous Agents and Multiagent Systems (AAMAS 2026)}{May 25 -- 29, 2026}
{Paphos, Cyprus}{C.~Amato, L.~Dennis, V.~Mascardi, J.~Thangarajah (eds.)}
\copyrightyear{2026}
\acmYear{2026}
\acmDOI{}
\acmPrice{}
\acmISBN{}

\setcopyright{none}
\renewcommand\footnotetextcopyrightpermission[1]{}

\acmSubmissionID{<<submission id>>}

\title[Population Synthesis]{Population synthesis with geographic coordinates}

\author{Jacopo Lenti}
\affiliation{
  \institution{Sapienza University of Rome}
  \city{Rome}
  \country{Italy.}
  \institution{CENTAI Institute}
  \city{Turin}
  \country{Italy}
  }
\email{jcp.lenti@gmail.com}

\author{Lorenzo Costantini}
\affiliation{
  \institution{CENTAI Institute}
  \city{Turin}
  \country{Italy}
  }
\email{lorenzo.costantini@centai.eu}

\author{Ariadna Fosch}
\affiliation{
  \institution{BIFI Institute, University of Zaragoza}
  \city{Zaragoza}
  \country{Spain.}
  \institution{CENTAI Institute}
  \city{Turin}
  \country{Italy}
  }
\email{arifosch@gmail.com}

\author{Anna Monticelli}
\affiliation{
  \institution{Intesa Sanpaolo Innovation Center}
  \city{Turin}
  \country{Italy}
  }
\email{anna.monticelli@intesasanpaolo.com}

\author{David Scala}
\affiliation{
  \institution{Intesa Sanpaolo}
  \city{Turin}
  \country{Italy}
  }
\email{david.scala@intesasanpaolo.com}

\author{Marco Pangallo}
\affiliation{
  \institution{CENTAI Institute}
  \city{Turin}
  \country{Italy}
  }
\email{marco.pangallo@gmail.com}

\usepackage[utf8]{inputenc} 
\usepackage[T1]{fontenc} 
\usepackage{amsmath,amsfonts}
\usepackage{array}
\usepackage[caption=false,font=normalsize,labelfont=sf,textfont=sf]{subfig}
\usepackage{textcomp}
\usepackage{stfloats}
\usepackage{url}

\usepackage{xcolor}
\usepackage{longtable}
\hypersetup{
bookmarks, pdftex,
colorlinks=true,
pagebackref=true, backref=page,
linkcolor={red!50!black},
filecolor={green!50!black},
citecolor={green!50!black},
urlcolor={blue!80!black},
}

\usepackage{verbatim}
\usepackage{comment}
\usepackage{balance}
\usepackage{xspace} 
\usepackage{graphicx} 
\usepackage{xfrac} 
\usepackage{mathtools} 
\usepackage{xr} 
\usepackage{cleveref} 
\usepackage{bbm}
\usepackage{dsfont}
\usepackage{placeins} 

\graphicspath{{../fig}}

\newcommand{\aamasspara}[1]{\noindent\textbf{#1}}

\newenvironment{squishlist}
{\begin{list}{$\bullet$}
 {\setlength{\itemsep}{0pt}
     \setlength{\parsep}{3pt}
     \setlength{\topsep}{3pt}
     \setlength{\partopsep}{0pt}
     \setlength{\leftmargin}{1.5em}
     \setlength{\labelwidth}{1em}
     \setlength{\labelsep}{0.5em} } }
{\end{list}}

\newcommand{\hidetext}[1]{}

\newcommand{\missref}[1]{\textbf{\textcolor{orange}{REF}}}
\newcommand{\VAE}{VAE\xspace}
\newcommand{\NF}{NF\xspace}
\newcommand{\NFs}{NF\xspace}
\newcommand{\enc}{\mathcal{E}\xspace}
\newcommand{\dec}{\mathcal{D}\xspace}
\newcommand{\copula}{copula\xspace}
\newcommand{\NFcopula}{NF+copula\xspace}
\newcommand{\shuffleCAP}{Local shuffle\xspace}
\newcommand{\shuffleprov}{Global shuffle\xspace}
\newcommand{\recloss}{\mathcal{L}_R\xspace}
\newcommand{\KLloss}{\mathcal{L}_{KL}\xspace}
\newcommand{\geoloss}{\mathcal{L}_{GEO}\xspace}
\newcommand{\NFVAE}{NF+VAE\xspace}
\newcommand{\datasetisp}{\texttt{data$\_$isp}\xspace}
\newcommand{\datasetairbnb}{\texttt{data$\_$airbnb}\xspace}
\newcommand{\sm}{SM\xspace}
\newcommand{\NhomesISP}{549247\xspace}
\newcommand{\NprovISP}{106\xspace}
\newcommand{\NfeaturesISP}{14\xspace}
\newcommand{\NnumfeaturesISP}{4\xspace}
\newcommand{\NcatfeaturesISP}{4\xspace}
\newcommand{\NboolfeaturesISP}{6\xspace}
\newcommand{\stepgrid}{0.01º\xspace}
\newcommand{\Ncityairbnb}{15\xspace}

\newcommand{\Ncatfeaturesairbnb}{1\xspace}
\newcommand{\Nnumfeaturesairbnb}{5\xspace}
\newcommand{\Nbinfeaturesairbnb}{7\xspace}
\newcommand{\Nintfeaturesairbnb}{4\xspace}
\newcommand{\MIAclassifier}{logistic regression\xspace}
\newcommand{\geosimNFVAE}{0.022\xspace}
\newcommand{\geosimNFcopula}{0.009\xspace}
\newcommand{\geosimVAE}{0.095\xspace}
\newcommand{\geosimshuffleCAP}{0.024\xspace}
\newcommand{\gridNFVAE}{0.391\xspace}
\newcommand{\gridNFcopula}{0.409\xspace}
\newcommand{\gridshuffleCAP}{0.410\xspace}
\newcommand{\moranNFVAE}{0.028\xspace}
\newcommand{\moranNFcopula}{0.080\xspace}
\newcommand{\moranshuffleCAP}{0.043\xspace}
\newcommand{\utilityNFVAE}{0.196\xspace}

\newcommand{\utilitycopula}{0.163\xspace}
\newcommand{\utilityNFcopula}{0.283\xspace}
\newcommand{\utilityshuffleprov}{0.206\xspace}
\newcommand{\zerocellNFVAE}{0.65\xspace}

\begin{abstract}
It is increasingly important to generate synthetic populations with explicit coordinates rather than coarse geographic areas, yet no established methods exist to achieve this. 
One reason is that latitude and longitude differ from other continuous variables, exhibiting large empty spaces and highly uneven densities. 
To address this, we propose a population synthesis algorithm that first maps spatial coordinates into a more regular latent space using Normalizing Flows (NF), and then combines them with other features in a Variational Autoencoder (VAE) to generate synthetic populations.
This approach also learns the joint distribution between spatial and non-spatial features, exploiting spatial autocorrelations.
We demonstrate the method by generating synthetic homes with the same statistical properties of real homes in 121 datasets, corresponding to diverse geographies. 
We further propose an evaluation framework that measures both spatial accuracy and practical utility, while ensuring privacy preservation. 
Our results show that the NF+VAE architecture outperforms popular benchmarks, including copula-based methods and uniform allocation within geographic areas. 
The ability to generate geolocated synthetic populations at fine spatial resolution opens the door to applications requiring detailed geography, from household responses to floods, to epidemic spread, evacuation planning, and transport modeling.
\end{abstract}


\begin{document}

\pagestyle{fancy}
\fancyhead{}

\maketitle

\section{Introduction}
In data-driven Agent-Based Models (ABMs)~\cite{pangallo2024datadriven}, it is crucial to place agents or elements of the environment at specific geographic coordinates. 
For instance, in ABMs of household responses to flood risk, homes must be located at coordinates with varying exposure to inundation~\cite{filatova2015empirical,haer2017integrating,pangallo2024climate}. 
Similarly, in epidemiological ABMs, individuals need to be placed at real residences, schools, or workplaces to capture realistic patterns of disease spread~\cite{aleta2022quantifying,pangallo2023unequal}. 
Beyond these cases, geolocation is also critical in applications such as traffic simulation, urban mobility, and evacuation planning, where distance and spatial constraints shape interactions~\cite{heppenstall2011agent}. 
For all these examples, geographic aggregates such as postcode areas or administrative units are not sufficient.

In an ideal setting, spatial ABMs could be initialized directly from geolocated data. 
In practice, however, this is rarely feasible, often due to privacy concerns, since geolocation can easily reveal individual identities. 
This creates the need for methods that construct synthetic populations of individuals or of the places where they live and work, such as homes, schools, or workplaces. 
In some cases, spatial information may be available from GPS traces, land-use maps, or similar sources, in which case the task reduces to assigning synthetic agents to specific places—a process still typically guided by rules of thumb~\cite{chapuis2018gen, zhou2022creating}. 
More commonly, only a geolocated sample is available, without land use maps or similar spatial information. 
The sample may be in a secured server, and it may not be possible to directly use the data for ABM simulation. 
The challenge then is to generate geolocated synthetic populations that reproduce the observed geographic distribution---for instance, placing residential units in residential areas rather than in water bodies or industrial zones. 
To the best of our knowledge, no method currently exists to accomplish this.

\begin{figure*}
    \centering
    \includegraphics[width=0.95\linewidth]{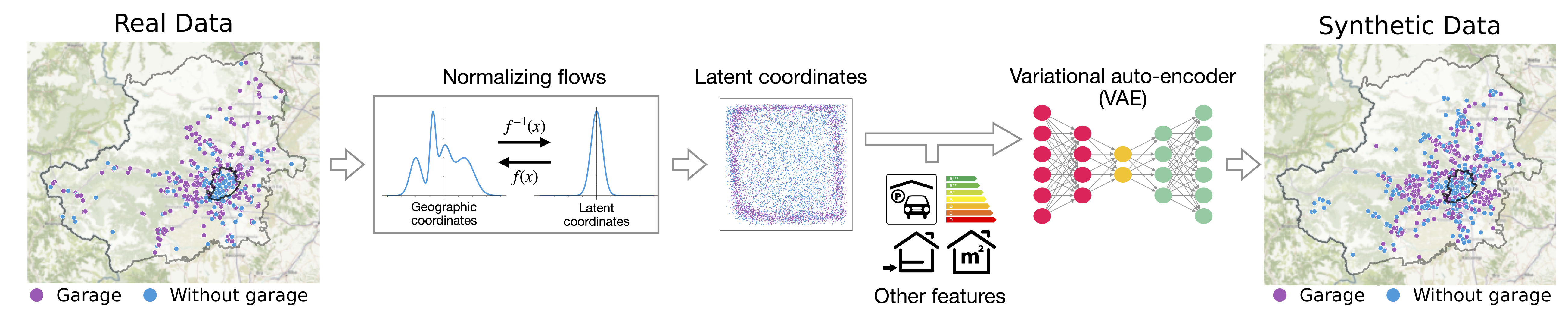}
    \caption{Overview of the proposed population synthesis generation approach.
    The real geolocated population is given as input to the generator.
    Normalizing Flows are trained to map the real geographic coordinates to a simple latent space.
    Together with all other home features, these latent coordinates are used to train a Variational Autoencoder.
    Finally the Variational Autoencoder samples synthetic populations that resemble the input data. 
    Left and right panels compare a random sample of 1,000 real and synthetic homes (respectively) in the province of Turin (gray lines). 
    Synthetic data reproduces real patterns, with higher presence of garage in the outskirts of Turin city (black lines in the maps), and lower presence in the city center.}\label{fig:map_TO}
    \Description{Overview of the proposed population synthesis.}
\end{figure*}

One reason why population synthesis methods have so far struggled to incorporate geographic coordinates is simply that it is difficult. 
Spatial data involve large empty areas and highly variable densities, and treating coordinates like standard continuous variables that have far more regular distributions is unlikely to yield meaningful results. 
The common alternative has been to use geographic aggregates (such as census tracts or provinces) as discrete variables~\cite{beckman1996creating,arentze2007creating,ye2009methodology,barthelemy2013synthetic,de2024gensynthpop}, but this approach is far from ideal, even beyond the issue that it may provide a too coarse grained spatial description. 
Using categorical variables for space is problematic partly due to the curse of dimensionality, and partly because it ignores spatial correlations. 
For example, when generating synthetic homes, a model may learn from the sample that in census tracts in the center of a historical city it is unlikely to find homes with a garage, yet this information does not transfer to adjacent tracts in the city, as a tract downtown is treated like a tract in the outskirts.

In this paper, we introduce a method for constructing synthetic populations with geographic coordinates, addressing both challenges outlined above in a unified framework. 
First, we tackle the non-regularity of spatial data by mapping coordinates into a latent space using Normalizing Flows (NF), a recently developed class of neural networks~\cite{papamakarios2021normalizing}. 
Next, we feed these latent coordinates, along with other features, into a Variational Autoencoder (VAE)~\cite{kingma2019introduction}, so that their joint representation in a second latent space captures correlations between spatial and non-spatial variables. 
As a generative model, this framework allows us to produce arbitrarily many synthetic samples that replicate both spatial distributions and feature relationships. \Cref{fig:map_TO} qualitatively shows the accuracy of our approach and provides a schematic representation of the generative model. Examples for other provinces and different attributes are available in Supplementary Materials (\sm).

Because this is the first paper generating a synthetic population with geographic coordinates, there are no off-the-shelf evaluation methods. 
Thus, a contribution of this paper is to propose an evaluation protocol that can be used by future methods that build on our work. 
Specifically, our protocol evaluates models along three dimensions: fidelity, utility, and privacy. 
Fidelity measures how closely the synthetic population resembles the real one. 
Utility assesses whether the synthetic data can be effectively used for real-world tasks. 
Privacy ensures that no sensitive information from the original individuals can be inferred or recovered from the synthetic population.

We find that our method (NF+VAE) beats other baselines at the combination of these metrics in the two case studies evaluated: (i) the generation of synthetic homes using a mortgage dataset in Italy~\cite{bellaver2025floods}; (ii) the generation of homes listed in Airbnb in \Ncityairbnb cities~\cite{airbnb, insideairbnb}.
In both cases, the inclusion of \NF is crucial to obtain an accurate mapping of the geographic coordinates, while the use of the \VAE allows us to capture spatial autocorrelation better than other traditional methods. 

Summarizing, \NFVAE  is an adaptable method that can be used to generate geolocated synthetic populations with a combination of numerical and categorical features. 
Moreover, due to the loss of information during the \VAE encoding, the synthetic population does not replicate individual records from the original dataset, thereby fully preserving privacy. 
Our approach opens the door to improved sharing of geolocated sensitive data, by producing synthetic datasets that remain faithful to the real ones, while safeguarding user privacy.

\aamasspara{Example.}
Our main case study concerns synthetic homes used as mortgage collateral. 
In collaboration with Intesa Sanpaolo (one of the leading commercial banks in Italy) and Intesa Sanpaolo Innovation Center, we developed an ABM to study the impact of flood risk on housing prices, as households shift their demand from at-risk homes to safer ones. 
For this application, it was essential that the geographic distribution of homes in the simulation reflected actual flood risk maps: it would make little sense to place most homes in high-risk zones if only a small fraction are located there in reality. 
At the same time, spatial information is highly sensitive, since precise coordinates could make it possible to identify individual clients. These two conditions illustrate a typical use case for our model. 
By building synthetic populations of homes, we can recreate geographic distributions present real data, while avoiding privacy leakage of sensible information about the bank's clients.
Finally, the synthetic populations can be used as input of the ABM simulations, thus maintaining the properties of the original dataset and preserving privacy.

\section{Problem statement}
We consider a dataset $D$ of size $N^R$, consisting of variables $x_1, \ldots,x_n$.
Among them, $x_1$ and $x_2$ correspond to latitude and longitude, while the remaining features may be numeric, integer, categorical, or boolean.
Our goal is to develop a generative model $\mathcal{G}$ that produces a synthetic dataset $\tilde{D} \coloneqq \mathcal{G}(D, N^S)$ of size $N^S$ with the same variables of $D$.
Specifically, $\tilde{D}$ should satisfy the following properties:
\begin{squishlist}
    \item Fidelity. Synthetic data should be similar to the original data. For a distance function $d$, we require $d(D, \tilde{D})$ to be small. The distance function can be defined in multiple ways, depending on whether the emphasis is on spatial coordinates, other features, or their combination.
    \item Utility. Synthetic data should be useful to draw realistic conclusions about the original data. If $m_D$ denotes a model trained on $D$, and $m_{\tilde{D}}$ the same model trained on $\tilde{D}$, then their output on a common input should be similar. Formally, for a distance $d$ defined on the output space of $m$, we require $d(m_D(D), m_{\tilde{D}}(D))$ to be small.
    \item Privacy. Synthetic data should not reveal sensitive information about individuals in $D$. We partition $D$ in $D^1$ and $D^2$. $D^1$ is used to train the generative model, i.e. $\tilde{D}^1 \coloneqq \mathcal{G}(D^1, N^S)$. Given a distance function returning the minimal distance between a sample $p$ and the elements of a population, privacy requires that a classifier cannot determine whether $p$ was used in the training of the generative model. In other terms, considering $p_1\in D_1$ and $p_2 \in D_2$, the distances $d(p_1, \tilde{D}^1)$ and $d(p_2, \tilde{D}^1)$ do not help a classifier detecting if $p_1\in D_1$.  This approach is in line with the Membership Inference Attacks (MIA)~\cite{shokri2017membership,steier2025synthetic,cormode2025synthetic} and evaluates if the generative model unintentionally exposes sensitive information contained in the real data.
\end{squishlist}
To address this problem, we first formalize the design of $\mathcal{G}$.
Afterwards, we propose meaningful distance measures and predictive models that operationalize fidelity, utility, and privacy.
Finally, we compare $\mathcal{G}$ against competitive generative approaches across these evaluation criteria.

\section{Methods}

We develop a model that takes as input a dataset of units, each associated with spatial coordinates. 
These units may represent homes, as in our case studies, or any type of geolocated entity. 
Each unit can be described by categorical, real, or discrete features.
The output of the model is a synthetic dataset of variable size that reproduces the joint distributions of the features with realistic spatial pattern, without replicating the original data.
    
\subsection{Preliminaries}
\label{sec:preliminaries}

Our proposal \NFVAE is a generative model that combines Normalizing Flows (\NFs) and Variational Autoencoders (\VAE).
In this section, we introduce these two frameworks.
We point to \cite{papamakarios2021normalizing} and \cite{kingma2019introduction, doersch2016tutorial} for more extensive reviews in the topics.

\aamasspara{Normalizing Flows.}
Normalizing Flows (\NFs) are a class of models that transform a simple distribution $Z$ into a more complex distribution $X$ through a sequence of $K$ invertible mappings $f_{i}$, each parameterized by $\theta$. 
Their invertible structure enables efficient mapping in both directions, from $Z$ to $X$ and vice versa.
The overall transformation can be written as $x = f_\theta(z) =  f_K \circ \ldots \circ f_2\circ f_1$~\cite{papamakarios2021normalizing}, where $z \sim Z$.
Since $f_i$ is bijective, it is possible to use the change of variable formula,

\begin{equation}
p_\theta(x) = p_\theta(z) \prod\limits_{i = 1}^K \left| \text{det} \left( \frac{\partial f_i ^{-1}}{\partial z_i} \right) \right| = p_\theta(z) \left| \text{det} \left( \frac{\partial f ^{-1}}{\partial x} \right) \right|,
\label{eq:change_variable}
\end{equation}
where $z_i = f_i(z_{i-1})$.
In practice, $f_i$ are chosen so that the Jacobian determinants in \Cref{eq:change_variable} are efficient to compute, ensuring tractable log-likelihoods.
Training proceeds by minimizing the KL divergence between the data distribution and the flow-based model.
In such a way, the model learns $\theta$ and, consequently, the bijective mapping between the base and data distributions.
The forward pass maps data into the simple base space, while the backward pass maps samples from the base back to the real data distribution.
\NFs are widely adopted with different goals. 
In generative modeling, they allow to generate new samples by drawing $z \sim Z$ and applying $f_\theta(z)$, while in variational inference they provide expressive approximate posteriors when the true posterior is unknown.
In our settings, Normalizing Flows represent flexible frameworks for capturing complex geographic patterns in a simple latent space.
A wide array of flow architectures have been proposed, including Planar and Radial Flows, Autoregressive Flows, Piecewise Linear and Piecewise Quadratic Flows~\cite{kobyzev2020normalizing}. 
In this work, we adopt Neural Spline Flows~\cite{durkan2019neural}, which are invertible transformations based on monotonic rational-quadratic splines, offering both flexibility and computational tractability.

\aamasspara{Variational Autoencoders.}
Variational Autoencoders (\VAE) are another class of generative models widely adopted in machine learning~\cite{kingma2019introduction}.
\VAE rely on two components, an encoder $\mathcal{E}$ and a decoder $\dec$.
The encoder $\enc$ maps the original data distribution $X$ into a lower-dimensional latent space with a simple prior distribution (typically Gaussian), $Z = \enc(X)$.
The decoder $\dec$ reconstructs the samples from the latent space back into the data space, $\tilde{X} = \dec(Z)$.
Both $\enc$ and $\dec$ are neural networks, whose parameters are learned during a training phase.
These neural networks are trained by minimizing a loss function that is composed of the reconstruction loss $\recloss$ and the KL-divergence $\KLloss$.
On the one hand, $\recloss$ measures the discrepancy between $X$ and $\tilde{X}$ to ensure the generated samples resemble the original.
On the other hand, $\KLloss$ regularizes the latent representation by aligning $Z$ with the chosen prior distribution.
Since the latent space has dimension that is lower than the data space, the encoding of $\enc(X)$ implies a discard of information.
As a result, the decoder learns to generate new samples that are similar to $X$, but not identical.

\subsection{Generative Model}
\label{sec:model}

Our proposed model, \NFVAE, generates synthetic geolocated data points by combining \NFs and \VAE.
Let $D$ denote the target $n$-dimensional dataset, and $D_{(x,y)}\in \mathbb{R}^2$ the subset of $D$ containing only the geographic coordinates.
We first train a \NF that maps $D_{(x,y)}$ into a simpler latent distribution, $Q^{NF} \in \mathbb{R}^2$.
This step is crucial, because geographic coordinates usually encode highly complex spatial patterns associated with natural and artificial constraints, such as mountains, lakes, urban barriers, and areas with varying population densities.
Thus, the \NF transforms $D_{(x,y)}$ into latent coordinates $Z_{(x,y)}$.

We then construct $D^{NF}$, which is a copy of $D$, where $D_{(x,y)}$ is replaced by $Z_{(x,y)}$.
Next, we train a \VAE to generate synthetic data resembling $D^{NF}$.
The encoder maps $D^{NF}$ into a $k$-dimensional Gaussian latent space, with $k < n$, while the decoder reconstructs the original space.
We define the loss of the VAE as $\mathcal{L}_{VAE} = \alpha_{GEO} \geoloss + \alpha_R \recloss + \alpha_{KL}\KLloss$, where (i) $\geoloss$ is the Euclidean distance between the geographic coordinates of the original and the synthetic data, (ii) $\recloss$ is the Euclidean distance between all other features, and (iii) $\KLloss$ is the KL-divergence between the encoded data and the Gaussian distribution.  
Since small errors of the geographic coordinates in the synthetic data lead to the generation of implausible samples, we decided to weight geographic features more than the others features, by setting $\alpha_{GEO} > \alpha_{R}$.
Given the granularity of our datasets, we are not interested in generating samples that differ too much from the available ones.
For this reason, we prioritize realistic reconstruction over excessive variation.
and $\alpha_R > \alpha_{KL}$.
Once $\alpha = (\alpha_{GEO}, \alpha_R, \alpha_{KL})$ and the hyperparameters of the neural networks  are chosen, we can train the model. 

In order to generate $N^S$ synthetic observations, we draw $N^S$ $k$-dimensional Gaussian samples and feed them to the decoder.
The trained decoder maps these samples to the output $n$-dimensional space, generating $\tilde{D}^{NF}$.
Therefore, we apply the trained \NF to map $\tilde{D}^{NF}_{(x,y)}$, which is distributed as $Q^{NF}$, back to the realistic geographic coordinates.
This yields to the final synthetic dataset $\tilde{D}$, containing $N^S$ observations and the same variables of $D$.
Thanks to the design of the loss, the generated samples are jointly distributed similar to the original data.
However, given the compression of information induced by the encoder of the \VAE, the generated samples are not identical to the original ones.

\subsection{Evaluation setup}
\label{sec:validation}

\begin{figure}[tb!]
    \centering
    \includegraphics[width=0.8\linewidth]{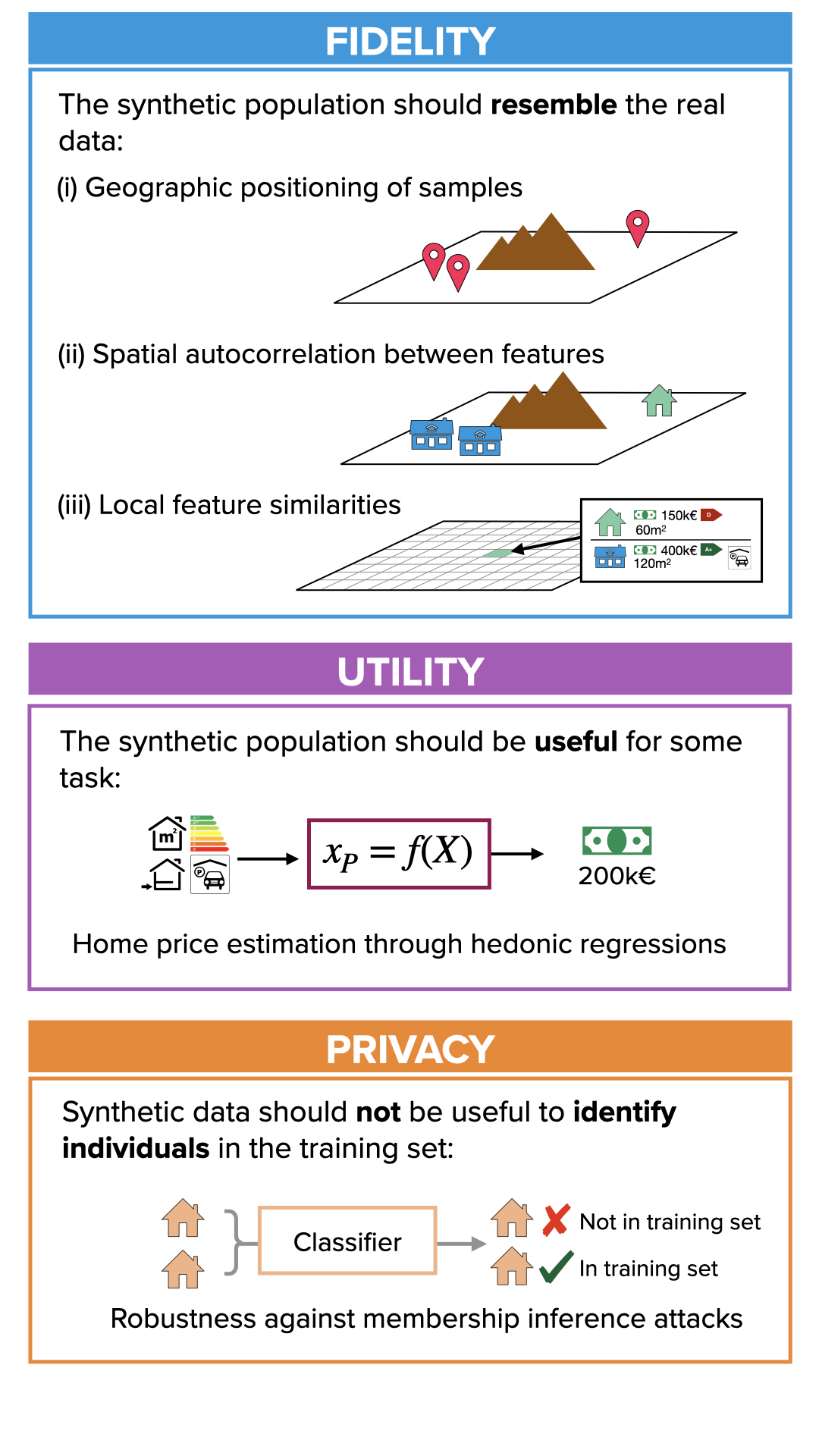}
    \caption{Description of the evaluation framework, based on fidelity, utility, and privacy.
    Fidelity measures (i) the similarity of the distribution of geographic coordinates, (ii) the similarity of the spatial autocorrelations, and (iii) the similarity of the houses generated in each grid cells.
    Utility assesses the quality of a model trained on synthetic data in predicting the house prices in real data.
    Privacy measures the robustness against membership inference attacks.    
    }
    \Description{Evaluation pipeline}
    \label{fig:pipeline}
\end{figure} 

We design an evaluation pipeline (see \Cref{fig:pipeline}) to compare the synthetic populations generated by our proposed method against a set of benchmark models.
Following prior works on synthetic data~\cite{cormode2025synthetic}, we evaluate the quality of synthetic data along three dimensions, which are \emph{fidelity}, \emph{utility}, and \emph{privacy}.
First, the synthetic data must closely resemble the original data, reproducing both distributional patterns and correlations among variables.
Given the inherent differences between geographic coordinates and the other features, we adopt three complementary measures of fidelity based on (i) geographic coordinates, (ii) spatial autocorrelation, and (iii) local features.
In \sm, we also assess the fidelity of correlations between non-geographic features.
Second, the synthetic data must be useful, which means that the analyses performed on them must yield insights comparable to those derived from the original data. 
In practice, this implies that models trained on synthetic data should provide results consistent with models trained on real data.
Third, synthetic data must preserve privacy.
Consequently, the synthetic data should not permit the recovery of individual information from the original dataset.

\aamasspara{Fidelity - Geographic coordinates.}
To assess the similarity of the distributions of geographic coordinates, we adopt the sliced-Wasserstein distance ($SW$)~\cite{nietert2022statistical}.
The Wasserstein distance, also known as Earth mover's distance, builds on Optimal Transport theory, and it quantifies the minimal cost of transforming one distribution into another.
Given two probability measures $\mu$ and $\nu$ on a metric space $(X, d)$, the $p-$Wasserstein distance is the minimal expected cost of transporting mass to transforms $\mu$ into $\nu$, where the cost is measured as the $p$-th power distance between points.
However, in high-dimensional settings, computing the Wasserstein distance becomes computationally expensive.
The sliced-Wasserstein distance mitigates the issue, by maintaining the properties of the classical Wasserstein distance, while significantly reducing the computational cost~\cite{kolouri2019generalized}.
The sliced-Wasserstein distance, $SW$, projects the high-dimensional distribution onto many one-dimensional subspaces and averages the Wasserstein distance between such projections~\cite{bonneel2015sliced}.
Relying on previous work, we set $p = 2$~\cite{flamary2021pot}.

The distance used to measure the fidelity of the geographic coordinates is 
\begin{equation}
\label{eq:fidelity_geo}
    d^F_{G} = SW\left(D_{(x,y)}, \tilde{D}_{(x,y)}\right).
\end{equation}

\aamasspara{Fidelity -  Spatial autocorrelation.}
Beyond geographic coordinates, synthetic data should preserve feature correlations, particularly spatial autocorrelations.
We measure this using Moran's Index~\cite{moran1948interpretation}, which measures the correlation between the spatial proximity and the similarity of a variable of interest $x$ for each pairs of data points. Spatial proximity is measured through a weight matrix $W = \left(w_{ij}\right)_{i,j = 1}^{N}$, which assigns higher weights to pairs of observations that are closer with each other.
In our study, we adopt a distance-based weight $w_{i,j} = \mathds{1}_{d(x,y) < m}$, where $d(i,j)$ is the Euclidean distance between the geographic coordinates and $m$ the first percentile of all pairwise distances $d(x,y)$.
In \sm, we report a robustness analysis with an alternative weighting function.

The computation of spatial autocorrelation for high-dimensional data is an open problem, and we define a measure based on Principal Components (PCs).
From the real data, we remove the geographic coordinates and we extract the first PCs, $y_1,\ldots,y_l$, explaining $95\%$ of variance, then project the synthetic data onto the same subspace.
For each $y_j$ we compute Moran's Index for the real data, $I_j$, and for the synthetic data, $\tilde{I}_j$.
We then compute Moran's Index of real data as $I=\sum_{j = 1}^l \lambda_j I_j$, where $\lambda_j$ is the eigenvalue associated with the $j$-th PC.
Analogously, $\tilde{I}=\sum_{j = 1}^l \lambda_j \tilde{I}_j$ is the Moran's Index of the synthetic data.
The distance measure used to quantify the fidelity of the spatial autocorrelation is
\begin{equation}
    \label{eq:fidelity_spatial}
    d^F_{S} = \left| I - \tilde{I} \right|.
\end{equation}
This definition of $d^F_{S}$ allows to reduce the number of features, while assigning more weight to the most explanatory components.

\aamasspara{Fidelity - Local features.}
Subsequently, we evaluate the fidelity of the features at a local scale. 
Similar to previous work on synthetic tabular data~\cite{herurkar2025evaluating}, we lay on Principal Component Analysis (PCA) to evaluate the fidelity of the generative model.
As done for computing $d^F_S$, we remove latitude and longitude, we extract the PCs $y_1,\ldots,y_l$ from the real data explaining $95\%$ of variance, and then we project the synthetic data onto the same subspace.
Following methods from climate science and geography~\cite{auffhammer2014empirical,kimerling1999comparing}, we partition the study region into a uniform grid of step size \stepgrid (approximately 1km).
For each cell $c_h$ in the grid, we compute $y^{h}_j$ and $\tilde{y}^{h}_j$, which are the projections on $y_j$ of the real and synthetic datasets, after filtering only the observations falling within $c_h$.
We compute $d_{L,h}^F = \sum\limits_{j = 1}^l\lambda_j\left(y^{h}_j - \tilde{y}^{h}_j\right)^2$, where $\lambda_j$ is the eigenvalue associated with the $j$-th PC.
Finally, we average $d_{L,h}^F$ across the set of grid cells $C'$, which are the cells containing more than zero units both in real and synthetic data.
\begin{equation}
    \label{eq:fidelty_local}
    d^F_L = \frac{1}{\left| C' \right|}\sum\limits_{h: c_h \in C'} d_{L,h}^F
\end{equation}

In this way, we measure the fidelity of the synthetic data as the average distance between the average real observation in a grid cell and the average synthetic observation in the same cell.
By computing $d^F_L$ only on $C'$, we exclude all empty cells in any of the two datasets.
Our motivations for following this choice are that (i) cells that are not in $C'$ are not useful to evaluate the feature similarity between the two datasets, and (ii) spatial differences across the dataset are already measured in $d^F_G$ 

\aamasspara{Utility.}
In order to assess the utility of the synthetic populations, we want to quantify the discrepancy between the output of a model trained on real data and the output of the same model trained on synthetic data.
This approach is also known as Train on Synthetic Test on Real (TSTR)~\cite{esteban2017real}.

Relying on a typical real estate context, we develop a hedonic regression model that estimates the home price from the other features~\cite{rosen1974hedonic}. 
In the hedonic regression, we include the spatial fixed effects, which quantify the effect of being in a specific subregion on the home price. 
To account for this fixed effect, we include the categorical variable $x_{s}$ encoding the subregion associated with each observations, among the $K$ available subregion. 

Let $x_p$ be the home price, then we write the hedonic regression task for home $i$ as
\begin{equation}
    \log x_p^i = \sum_{j\neq p} \omega_j x_j^i + \sum_r\mu_r \mathds{1}_{x^i_s = l} +\epsilon^i,
    \label{eq:hedonic_regression}
\end{equation}
where $\mu_r$ is the spatial fixed effect related to subregion $r$, $\epsilon^i$ is the error term, and $\omega_j$ the regression coefficient for variable $j$.
We call $m_D$ the model~\eqref{eq:hedonic_regression} trained on the real data $D$, while $m_{\tilde{D}}$ denotes the model trained on the synthetic data $\tilde{D}$.

To assess the utility of a synthetic population, we use $m_{D}$ and $m_{\tilde{D}}$ to predict the home prices on $D$, and then we compare the performances of the two models.
To this end, we denote with $m_D(D)$ the home prices predicted by $m_D$ and $m_{\tilde{D}}(D)$ the home prices predicted by $m_{\tilde{D}}$.
Subsequently, we compute $R^2(m_D) \coloneqq R^2(m_D(D), x^p)$ and $R^2(m_{\tilde{D}}) \coloneqq R^2(m_{\tilde{D}}(D), x^p)$, where $R^2$ is the coefficient of determination.
Finally, the distance measure used to quantify the utility of the synthetic data is 
\begin{equation}
    \label{eq:utility}
    d^U = \lvert R^2(m_D) - R^2(m_{\tilde{D}}) \rvert .
\end{equation}
If $d^U$ is close to zero, this means that the synthetic data are as useful as real data to accomplish the task of predicting home prices with hedonic regression with spatial fixed effects.

\aamasspara{Privacy.} 
Many different measures have been proposed to quantitatively assess the privacy of a synthetic dataset~\cite{cormode2025synthetic}.
The general goal is to ensure that synthetic data cannot be used to recover sensitive information about individuals in the original dataset.
In this work we focus on the risk of membership inference~\cite{steier2025synthetic,shokri2017membership}, where an adversary attempts to determine whether a particular record was used to train the generative model.
Such attacks are concerning because, in some cases, the presence of a record in the training set may itself reveal sensitive information.
For example, in our settings, confirming that a home appears in the training implies that the bank issued a mortgage to the homeowner.

To quantify this risk, we design the following procedure.
First, we split $D$ into two disjoint datasets $D^1$ and $D^2$, where $D^1$ is a random sample of $95\%$ of $D$.
Second, we generate a synthetic population $\tilde{D}^{1}$ by training the generative model on $D^{1}$.
For any sample $p\in D$, we define $d(p, \tilde{D}^1)$ as the minimum Euclidean distance between $p$ and the closest $y\in \tilde{D}^1$. 
The distance considers all spatial and non-spatial features, after standard one-hot encoding and rescaling within $[0,1]$.
Third, we construct two disjoint subsets, $Z^{TRAIN}$ and $Z^{TEST}$, by splitting $D$ with a ratio 80-20.
Using $Z^{TRAIN}$, we train a classifier $\mathcal{C}$ that predicts whether a record $p$ belongs to $D^1$ based only on $d(p, \tilde{D}^1)$.
Fourth, we evaluate $\mathcal{C}$ on $Z^{TEST}$ using Area Under ROC Curve (AUC-ROC).
The AUC-ROC is commonly used in imbalanced datasets, and it measures the probability that a classifier assigns to a random positive a score higher than a random negative.
Thus, the privacy preservation of the synthetic population is measured as 
\begin{equation}
    \label{eq:privacy}
    \rho^P = \text{AUC-ROC}(\mathcal{C}(Z^{TEST})) -0.5.
\end{equation}
In this way, if $\rho^P > 0$, $\mathcal{C}$ may infer whether a record was part of the training dataset better than random classifier.
Since this classification is done by simply matching the record with the most similar observation in the synthetic dataset, this definition strongly relates to the notion of privacy.

We considered a \MIAclassifier as the classifier $\mathcal{C}$ in the main text. In \sm, we show that the results are robust considering other classifiers.

\section{Experimental Settings}
\label{sec:experiments}

\aamasspara{Data description.}
We evaluated the generative model using two buckets of dataset, \datasetisp and \datasetairbnb.

\datasetisp is a unique dataset owned by Intesa Sanpaolo (ISP), one of the leading commercial banks in Italy. 
Specifically, we got access to the data about the mortgages issued by Intesa Sanpaolo between January 2016 and August 2024 (all data were treated with a GDPR-compliant and privacy preserving protocol). 
These dataset collects information about the mortgages that have to be repaid up to August 2024. 
For each mortgage, we have information regarding the home given as collateral that is the input of the generative model since we have access to a suitable set of home-related features. 
The dataset contains \NfeaturesISP features, \NnumfeaturesISP features are numeric, \NboolfeaturesISP are boolean, and the remaining \NcatfeaturesISP are categorical. 
The numerical features include latitude and longitude of the home given as collateral, surface, and sale price, while energy class, cadastral code, floor, and construction year are represented as categorical features.
Boolean features include presence of air conditioning, annex (i.e., basement or rooftop storage room), and  garage among others.
The dataset contains \NhomesISP homes across \NprovISP Italian provinces. 
We built one dataset for each province containing the homes in the province at hand. 
7 provinces have $<1,000$ homes, 74 provinces have between $1,000$ and $5,000$ homes, 11 provinces range between $5,000$ and $10,000$, and 14 provinces have $>10,000$ homes.
In \sm, we provide a summary description of all the features in \datasetisp.

Secondly, we evaluate the method on \datasetairbnb, which lay on a public dataset of Airbnb provided by Inside Airbnb~\cite{insideairbnb}.
Airbnb is a widely used online marketplace for short-term rentals, where hosts describe their properties through a variety of attributes such as location, number of rooms, available amenities, and price per night.
We selected a sample of \Ncityairbnb cities across different countries and, for each city, we created a dataset comprising  \Ncatfeaturesairbnb categorical, \Nbinfeaturesairbnb binary, \Nintfeaturesairbnb discrete, and \Nnumfeaturesairbnb continuous features.
A detailed description of \datasetairbnb is provided in \sm.

\aamasspara{Benchmarks.}
We compare our proposed method, \textbf{\NFVAE} (see \sm for implementation details), with a set of benchmarks of particular interest.
\begin{squishlist}
    \item \textbf{\VAE}. As an ablation study, we generated synthetic data with \VAE, without transforming the geographic coordinates. In this case, we use the same \VAE architecture of \NFVAE.
    \item \textbf{Copula}. We generate synthetic populations with Gaussian copula, relying on the implementation of \texttt{sdv}~\cite{patki2016synthetic}.
    This method fits a multivariate Gaussian distribution in a latent space, then uses marginal inverse-CDF transformations to map latent samples into the original data domain. 
    In such a way, it maintains marginal distributions while approximating the correlations through the Gaussian copula~\cite{nelsen2006introduction}.
    \item \textbf{\NFcopula}. We transform the geographic coordinates with the same \NFs used in \NFVAE, but replacing \VAE with Gaussian copula.
    In such a way, we evaluate also the contribution of \NFs applied on a different generative model.
    \item \textbf{\shuffleprov}. This method samples the original observations with replacement and then assigns geographic coordinates uniformly random within the region covered by the dataset. In our case, the regions of the synthetic populations are the Italian provinces (\datasetisp) and the cities (\datasetairbnb).
    
    \item \textbf{\shuffleCAP}. Similar to the previous method but it samples with replacement the observations in the original dataset, while assigning random coordinates within the boundaries of the same granular subregion of the sampled observation. The subregions in \datasetisp are the postcode areas, while the subregions of \datasetairbnb are the neighborhoods.
\end{squishlist}

\begin{figure}[tb]
    \centering
    \includegraphics[width=0.85\linewidth]{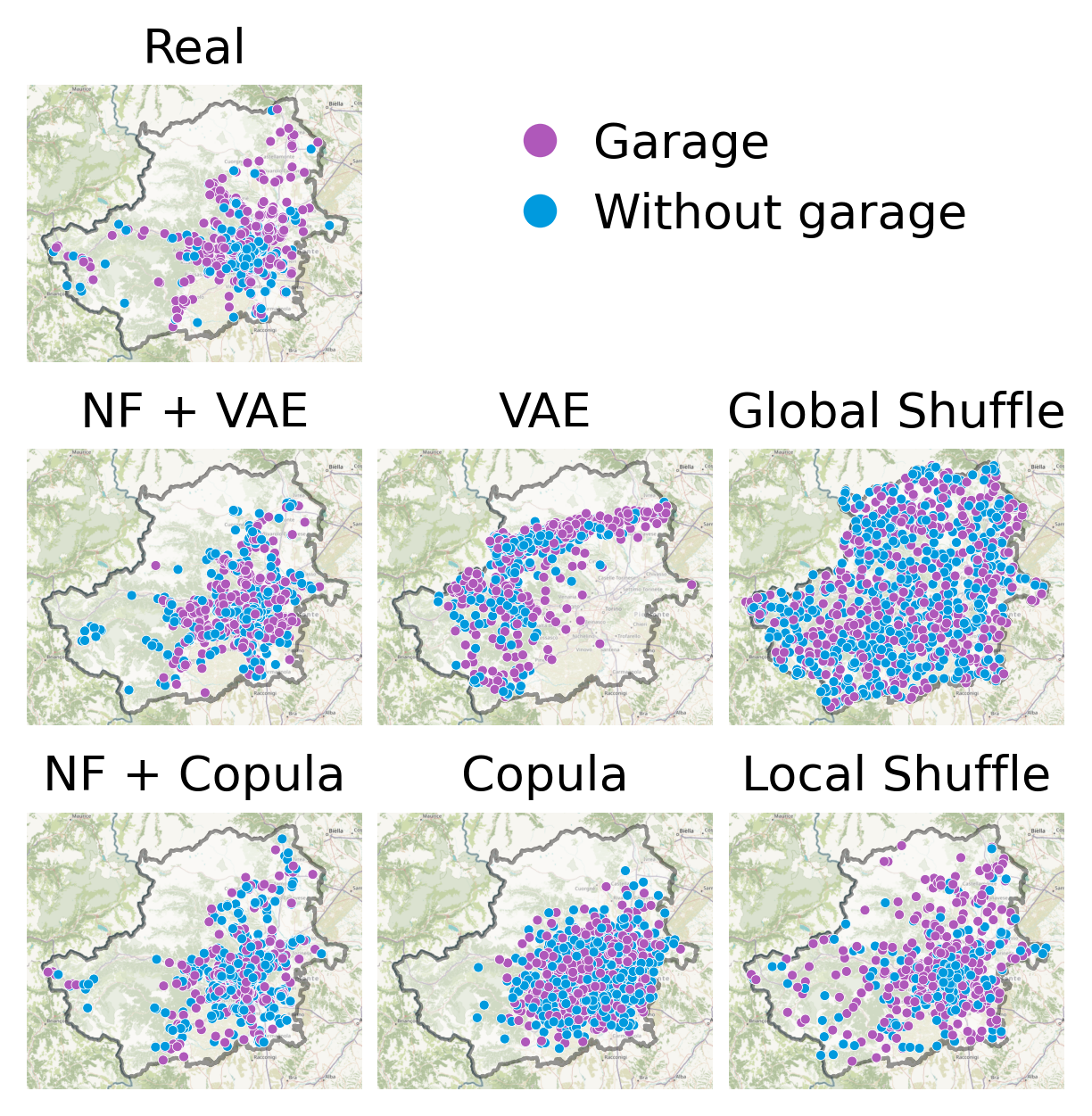}
    \caption{Real and synthetic homes generated by the benchmark generators in the province of Turin.
    In this plot, for each map we show a random sample of 1,000 homes, colored by the presence of garage.
    }
    \Description{Maps all baselines}
    \label{fig:all_maps_TO}
\end{figure}

By choosing these benchmarks, we compare \NFVAE with a natural competitor (\NFcopula), two ablated models (\VAE and \copula), and two non-trivial null models (\shuffleprov and \shuffleCAP).
For all benchmarks, we filter out the synthetic data points that fall outside the borders of the region by matching them against the region's shapefile.
In \Cref{fig:all_maps_TO}, we show a visual representation of a synthetic population for each method.
 
\section{Results}
\label{sec:results}

\begin{figure*}[ht]
    \centering
    
    \includegraphics[width=\linewidth]{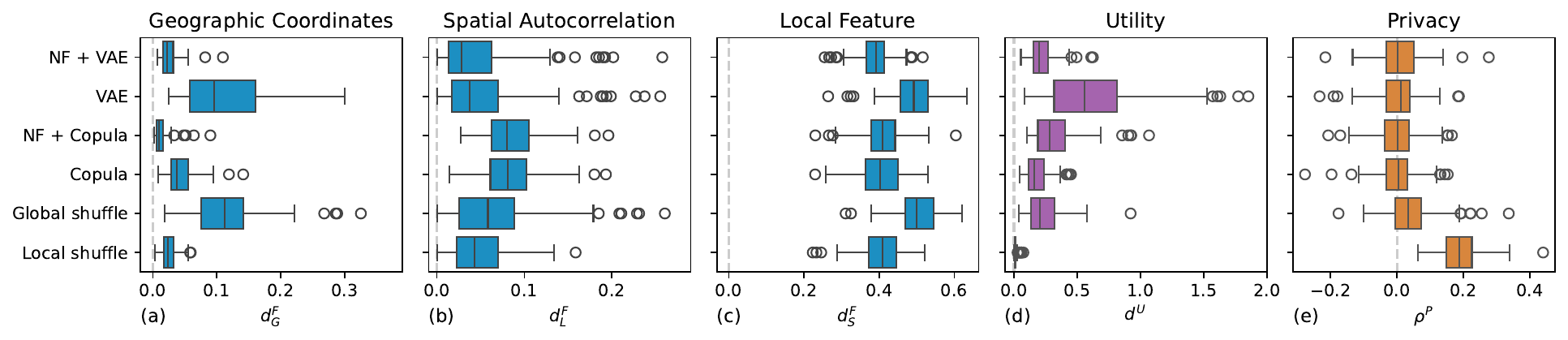}
    \caption{Distributions of evaluation metrics in \datasetisp.
    (a) \emph{Fidelity - Geographic coordinates}, i.e., sliced-Wasserstein distance geographic coordinates,
    (b) \emph{Similarity - Spatial autocorrelation}, distance between spatial autocorrelations in the PCs of real and synthetic homes,
    (c) \emph{Fidelity - Local features}, distance between average home per spatial grid cell,
    (d) \emph{Utility}, distance between $R^2$ in predicting real log-price with a model trained with real and synthetic data,
    (e) \emph{Privacy}, difference between AUC-ROC of a classifier trained to infer the membership in the original dataset.
    Best performances are close to 0 in all methods.
    (a), (b), (c), and (d) are always positive, (e) can be negative.
    Detailed statistics of this figure are available in \sm.
    }
    \Description{Boxplots evalutation}
    \label{fig:boxplots_results}
\end{figure*}

We evaluate model performance following the validation pipeline described in \Cref{sec:validation}, along the dimensions of fidelity, utility, and privacy. Our main results show the performance of the model when applied to \datasetisp, while we include an equivalent validation analysis in the \sm using \datasetairbnb. 

\aamasspara{Fidelity.}
First, we focus on the fidelity of geographic coordinates, by computing the sliced-Wasserstein distance with 1000 projections.
As shown in \Cref{fig:boxplots_results}a, the methods that use \NFs achieve performances comparable to \shuffleCAP (median $d^F_G$: \geosimNFVAE for \NFVAE, \geosimNFcopula for \NFcopula, and \geosimshuffleCAP for \shuffleCAP).
By contrast, \VAE fails to adequately capture the geographic distribution (median $d^F_G$: \geosimVAE).
This highlights the importance of transforming geographic coordinates through a trained \NF, before using them in the generative model.
Second, when evaluating the spatial autocorrelations (\Cref{fig:boxplots_results}b), we observe that the VAE-based methods capture spatial dependencies better than copula-based methods (median $d^F_S$: \moranNFVAE for \NFVAE, \moranNFcopula for \NFcopula, and \moranshuffleCAP for \shuffleCAP). 
In \sm we show that the results of \Cref{fig:boxplots_results}b are consistent by choosing a different weighting function in the computation of Moran's I. 
Third, in the local features fidelity (\Cref{fig:boxplots_results}c) \NFVAE and \NFcopula outperform the other methods (average distances: \gridNFVAE for \NFVAE, \gridNFcopula for \NFcopula, and \gridshuffleCAP for \shuffleCAP).
These results indicate that both methods generate plausible homes within individual grid cells.

\aamasspara{Utility.}
We use hedonic regressions, where the spatial fixed effects are determined by the postcode areas for \datasetisp, and the neighborhoods for \datasetairbnb.
Since \shuffleCAP replicates the exact same features of the original data, it achieves the highest utility.
However, when comparing the remaining methods, we observe that \NFVAE and \copula outperform \shuffleprov (median $d^u$: \utilityNFVAE for \NFVAE, \utilityNFcopula for \NFcopula, \utilitycopula for \copula, \utilityshuffleprov for \shuffleprov).
Since \shuffleprov merely replicates homes with random locations, the high utility achieved by \NFVAE (and \copula) supports the importance of correctly replicating the spatial distributions of samples in building reliable model.

\aamasspara{Privacy.}
Finally, we compare the privacy robustness against a Logistic Regression classifier for the analyzed benchmarks, \Cref{fig:boxplots_results}e.
We note that \shuffleCAP is significantly larger than 0, showing weak robustness against membership inference attacks.
This means that it is possible to classify a data point as part of the training set based on the closest synthetic data point. 
All other methods are robust against the classifier, as the privacy measure is not significantly different from zero.
In \sm we provide an extension of \Cref{fig:boxplots_results}e, showing that these privacy robustness is consistent across a set of standard machine learning classifiers.

\aamasspara{Summary.}
By looking at the overall results, we can detect the following insights.
\begin{squishlist}
\item The study of the ablated models reveal the importance of \NFs in learning the spatial distributions.
\item \VAE-based models capture the spatial autocorrelations, thus favoring \NFVAE over \NFcopula.
\item \NFVAE also produces synthetic populations that are useful for downstream models.
\item All non-shuffle models are robust privacy attacks, thus excluding \shuffleCAP.
\end{squishlist}

Overall, our experimental analyses show the superiority of \NFVAE in the combination of fidelity, utility, and privacy.
In \sm we show that similar conclusions hold for \datasetairbnb.
However, in \datasetairbnb we also notice that \NFcopula achieves performances that are comparable to \NFVAE.

\section{Discussion}
\label{sec:discussion}
We proposed and evaluated a method for generating geolocated synthetic populations that combines Normalizing Flows (\NFs) and Variational Autoencoders (VAE).
Our evaluation pipeline is based on fidelity, utility, and privacy.
Using this pipeline, we assess the quality of our model, \NFVAE, across an extensive set of datasets (\NprovISP in \datasetisp + \Ncityairbnb in \datasetairbnb).
Overall, our approach improves other competitors in terms of fidelity and utility, while maintaining the privacy in the generation of geolocated units.

\noindent\emph{Flexibility.}
Our method requires minimal customization, beyond standard preprocessing and the definition of network architectures.
\NFVAE easily handles variables of different types and dimensions.
Additionally, it requires only the region shapefile, without the need of additional geographic information, such as subregion boundaries, streets, or points of interest.

\noindent\emph{Zero-cell problem.}
While traditional sampling-based methods struggle to reproduce combinations of features that are not present in the real data, \NFVAE generates novel but realistic combinations.
As shown in \sm, \shuffleCAP only replicates existing samples, thus failing to address zero-cell problem-
Contrarily, \NFVAE produces a median of \zerocellNFVAE homes that are not present in the real data.
This strength of characterizes this class of generative models: just as in computer vision \VAE produces realistic images not present in the training data, \NFVAE can synthesize homes that are plausible but not present in the original data. 

\noindent\emph{Scalability.}
Grounded in the wide framework of deep generative models, \NFVAE easily handles datasets with large sample sizes and high dimensionality.

\noindent\emph{Generalization.}
Although this work is motivated by housing market analysis, the method naturally generalizes to many other domains.
In ABMs, geolocated populations can represent diverse spatial entities, such as households at their residences, workers at their workplaces, or students at their schools. 
Since privacy concerns usually restrict the use of collected data, making synthetic data represents a compelling alternative.

\noindent\emph{Towards standard evaluation.}
To our knowledge, this is the first systematic comparison of synthetic population generators for geolocated data.
In accordance with the usual practices in machine learning, we underline the importance of rigorous evaluation of the proposed method, with a systematic benchmark against representative baselines.
We hope that this work fosters further research on the evaluation of geographic synthetic data.

\aamasspara{Limitations.} 
Despite these strengths, several limitations remain.

\noindent\emph{Model implementation.}
Since \NFVAE combines two deep learning approaches, it reflects a set of modeling and optimization choices.
Although we demonstrated that our design meets the initial requirements, refined implementations of \NF and \VAE  could further improve the quality of the generated data.

\noindent\emph{Evaluation metrics.}
Our evaluation framework involves a set of arbitrary choices related to the measures of fidelity, utility, and privacy.
The adopted definitions and metrics are not exhaustive, and they can vary depending on the application.
For example, we evaluated utility based on the regression of home prices, while alternative measures and regression models might be more appropriate in other domains. In the case of ABMs, a reasonable utility measure could be comparing output of the ABM simulations when using the real or synthetic populations.
Similarly, we focused on membership inference as a privacy attack, but alternatives could provide equally relevant insights.
For example, attribute inference attack aims to recover sensitive individual attributes from the synthetic data and partial knowledge of the real data.
Given the absence of a unique superior standard for privacy in synthetic data, we recognize that evaluation must be tailored to the intended use case.
Another possibility to address privacy is to use generative models relying on differential privacy. 
Thus, the level of privacy can be defined in advance. 
We leave the comparison of \NFVAE with such approaches to future work.

\noindent\emph{Evaluation desiderata.}
In addition to fidelity, utility, and privacy, the evaluation could include other features, such as efficiency and expressivity~\cite{cormode2025synthetic}.
Regarding efficiency, we acknowledge that training \NFVAE requires substantially more computational resources than simpler approaches, such as copula-based or shuffle-based generators.
Our results, however, show that replacing the \VAE with \copula maintains comparable performances and significantly reduces the training cost.
Still, \NF remains crucial for transforming geographic coordinates in a convenient representation.
As for expressivity, given the strong capabilities demonstrated by \VAE in other domains, such as image generation, we expect \NFVAE to exhibit a comparable expressive power in population synthesis.

\noindent\emph{Benchmarks.}
The inclusions of additional baselines, such as Combinatorial Optimization, Generative Adversarial Networks (GANs), or Bayesian Networks, would provide a more comprehensive assessment.
Our benchmarks comprises (i) a simple and valid alternative (\NFcopula), (ii) the ablated models to isolate the effect of \NFs, and (iii) the suffle-based models, acting as null models.

\noindent\emph{Data availability.}
Finally, the effectiveness of \NFVAE depends on data availability.
Both \VAE and \NF perform best with large, high-dimensional datasets, and may struggle with overfitting or weak pattern learning when data are limited.
However, given the increasing availability of granular real-world datasets, we believe our approach is applicable in a wide range of context.

\aamasspara{Related work.} 
Population synthesis typically combines a few principled methods with many problem-specific assumptions~\cite{adiga2015generating}. 
Our contribution falls within the class of sample-based methods~\cite{borysov2019generate,garrido2020prediction}, where a sample is available from which all relevant statistical distributions can be learned and new instances generated. 
This differs slightly from the classical population synthesis problem, where one typically has a sample at a higher geographic aggregation and marginal distributions at the desired aggregation, often addressed with Iterative Proportional Fitting~\cite{beckman1996creating}. Sample-free methods also exist~\cite{barthelemy2013synthetic,de2024gensynthpop}, and active research is comparing them to sample-based approaches~\cite{lenormand2013generating}. 

While several directions have been explored within population synthesis (e.g., matching individuals to households~\cite{arentze2007creating,ye2020iterative}), two are particularly relevant for our work.
One line of research concerns the use of generative modeling, and VAEs in particular, for population synthesis~\cite{borysov2019generate,garrido2020prediction,aemmer2022generative,blackthorn2024training,sane2025comprehensive}. 
A key advantage is that, being probabilistic rather than deterministic, VAEs never return exact copies of individuals from the training sample, thereby preventing identification. 
Probabilistic approaches were already available through Bayesian Networks~\cite{sun2015bayesian,zhou2022creating}, Gibbs Sampling~\cite{farooq2013simulation}, and related methods. 
However, deep learning approaches offer at least two advantages in population synthesis. 
First, it helps overcome the curse of dimensionality, since traditional methods scale poorly with the number of features, and categorical variables treated with one-hot encoding further increase the dimensionality.
Second, it addresses zero-cell problem, exploring novel plausible combinations of features.

Another line of research concerns geolocated synthetic populations, moving beyond geographic aggregates. 
This area has received much less attention. 
A common approach is to distribute units randomly within polygons such as neighborhoods or along lines such as roads, as in the SPEW package~\cite{gallagher2018spew}. 
When map information is available, for example land use maps or GIS features~\cite{chapuis2018gen,zhou2022creating}, it can be used to place agents more precisely according to heuristic rules, such as avoiding industrial areas or water bodies for households. 
Our work is complementary to this research: it enables the assignment of geographic coordinates to agents in a principled manner even when map information is not available, provided that a sample with localization data exists.

In the wide context of synthetic data, the evaluation of the generated samples is an active field of research.
Synthetic data are used for different purposes, such as data sharing~\cite{assefa2020generating}, improvement of downstream models~\cite{dina2022effect}, and  fairness~\cite{van2021decaf}.
As synthetic data are required to be similar to real data, univariate and multivariate distance measures are employed~\cite{chapuis2018gen, hernandez2022synthetic, espinosa2023quality}.
Along with fidelity, privacy is a major concern, as many measures can be defined with different assumptions on the data or model availability to the attacker~\cite{osorio2024privacy,cormode2025synthetic,carvalho2022survey}.
As in our work, utility is a common choice for assessing the quality of the synthetic data~\cite{snoke2018general,hansen2023reimagining,hernandez2022synthetic}.
Additionally, synthetic data are evaluated in terms of expressivity, efficiency, diversity, and generalization~\cite{alaa2022faithful, cormode2025synthetic}.

\aamasspara{Reproducibility.}
All code used for model training and evaluation is available at \url{https://anonymous.4open.science/r/NFVAE-population-synthesis-CB06/}.
Due to strict privacy regulations governing financial data, we cannot release \datasetisp either at the individual or aggregate level.
However, all analyses conducted on the public available \datasetairbnb are fully reproducible.

\clearpage

\bibliographystyle{ACM-Reference-Format} 
\bibliography{AAMAS/bibliography}

\end{document}